\documentclass[11pt]{article}

\usepackage[preprint]{acl}

\usepackage{array}
\usepackage{booktabs}
\usepackage{makecell}
\usepackage{multirow}
\usepackage[table]{xcolor}
\usepackage{tabularx}
\usepackage{colortbl}
\usepackage{times}
\usepackage{latexsym}

\usepackage[T1]{fontenc}

\usepackage[utf8]{inputenc}

\usepackage{microtype}

\usepackage{amsmath}
\usepackage{amssymb}
\usepackage{inconsolata}
\usepackage{multirow}
\usepackage{booktabs}

\usepackage{graphicx}

\title{HISR: Hindsight Information Modulated Segmental Process Rewards \\ For Multi-turn Agentic Reinforcement Learning}

\author{
\normalsize
 Zhicong Lu\(^{1,2,3}\)\footnotemark[3], Zichuan Lin \(^{3}\)\footnotemark[3], Wei Jia\(^{1,2}\), Changyuan Tian\(^{1,2}\)
 \\
\normalsize
\textbf{Deheng Ye\(^{3}\), Peiguang Li\(^{1,2}\), Li Jin\(^{1,2}\), Nayu Liu\(^{4}\)\footnotemark[2], Guangluan Xu\(^{1,2}\), Wei Feng\(^{4}\)} 
 \\ 
 \vspace{-3px}
\small
\(^1\) Aerospace Information Research Institute, Chinese Academy of Sciences \\
 \vspace{-3px}
\small
\(^2\) University of Chinese Academy of Sciences, 
\(^3\) Tencent Hunyuan\\ 
 \vspace{-3px}
 \small
\(^4\) School of Computer Science and Technology, Tianjin University \\
 \vspace{-3px}
\small
 \text{nazaritelzc}@gmail.com, \text{liunayu.lny}@qq.com
}

\begin{document}
\maketitle
\renewcommand{\thefootnote}{\fnsymbol{footnote}}
\footnote{\textbf{Work on progress}}
\footnotetext[3]{Equal Contribution}
\footnotetext[2]{Corresponding author}
\begin{abstract}
While large language models excel in diverse domains, their performance on complex long-horizon agentic decision-making tasks remains limited. Most existing methods concentrate on designing effective reward models (RMs) to advance performance via multi-turn reinforcement learning. However, they suffer from delayed propagation in sparse outcome rewards and unreliable credit assignment with potentially overly fine-grained and unfocused turn-level process rewards. In this paper, we propose (\textbf{HISR}) exploiting \textbf{H}indsight \textbf{I}nformation to modulate \textbf{S}egmental process \textbf{R}ewards, which closely aligns rewards with sub-goals and underscores significant segments to enhance the reliability of credit assignment. Specifically, a segment-level process RM is presented to assign rewards for each sub-goal in the task, avoiding excessively granular allocation to turns. To emphasize significant segments in the trajectory, a hindsight model is devised to reflect the preference of performing a certain action after knowing the trajectory outcome. With this characteristic, we design the ratios of sequence likelihoods between hindsight and policy model to measure action importance. The ratios are subsequently employed to aggregate segment importance scores, which in turn modulate segmental process rewards, enhancing credit assignment reliability. Extensive experimental results on three publicly benchmarks demonstrate the validity of our method.
\end{abstract}

\section{Introduction}

Large language models (LLMs) have exhibited exceptional performance across diverse domains \cite{lu-etal-2024-rethinking, lu-etal-2025-piper, liu-etal-2025-sara, tian2025rectify}, even surpassing human performance in certain scenarios. However, their capabilities remain limited in complex long-horizon agentic decision-making tasks (e.g., household assistance \cite{shridhar2020alfworld, puig2018virtualhome}), where LLMs act as an agent to achieve multiple sub-goals in the task. To develop LLMs into comprehensive general artificial intelligence, it is crucial to explore effective solutions for agentic capabilities enhancement.

\begin{figure*}[http]
    \centering
    \includegraphics[width=1\linewidth]{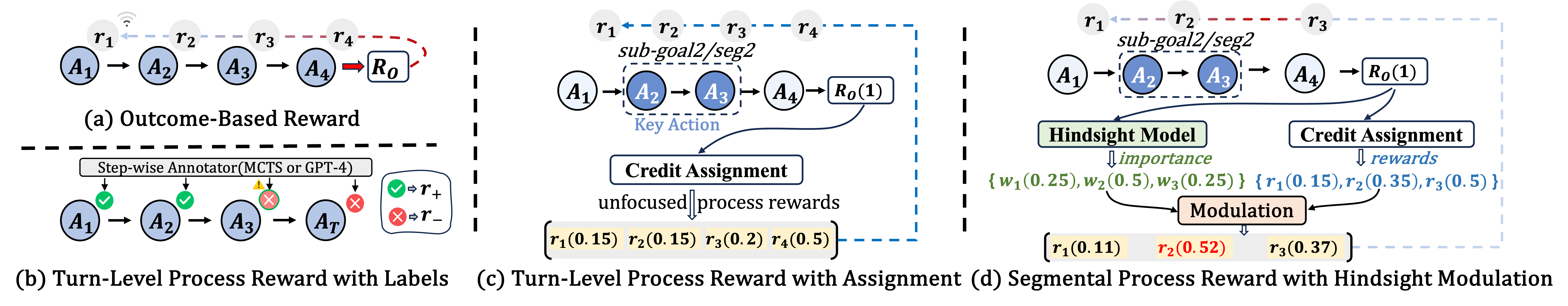}
    \caption{Comparison between mainstream reward models and our method in multi-turn reinforcement learning. $A_i$, $R_0$, $r_i$ denote the decided action in $i$-th turn, the outcome of trajectory, and process rewards. (b) and (c) assign rewards at turn-level granularity. The unfocused process rewards in (c) denote that credit assignment fails to consider the action importance, while ours (d) aligns rewards with sub-goals and underscores significant segments.}
    \label{fig_introduction}
\end{figure*}

To achieve this goal, most existing methods concentrate on designing effective reward models (RMs) to advance performance via multi-turn reinforcement learning (RL) algorithms (e.g., Proximal Policy Optimization, PPO \cite{schulman2017proximal}). The mainstream RMs can be broadly categorized into two paradigms: 1) \textbf{the outcome-based RM} \cite{ouyang2022training, shao2024deepseekmath} in Fig \ref{fig_introduction}a. merely assigns a single reward at the end of trajectory according to the trajectory outcome. However, in the agentic tasks, the exacerbation of long-horizon and the characteristic of delayed reward make it more difficult to propagate the final reward to actions that occurred earlier, struggling to guide the optimization of decision-making in the environment \cite{wang2025spa, jia2025u}.

To alleviate this issue, 2) \textbf{the turn-level process RM} is presented to perform the credit assignment within the trajectory, which can be roughly divided into two categories based on the requirement of process reward labeling. One (Fig \ref{fig_introduction}b.) utilizes finite Monte Carlo Tree Search (MCTS) \cite{wang2024math, zhang2025lessons} or GPT-4 \cite{achiam2023gpt, lightman2023let, xie2025capo, bi2025f2rvlm} to annotate the pseudo process rewards, yielding the noisy turn-level supervision. The others (Fig \ref{fig_introduction}c.) are independent of process reward labels, instead leveraging the outcome of the trajectory to indirectly monitor the credit assignment \cite{wang2025spa, yu2024steptool}.  Nevertheless, the former faces the difficulty of cost-effectively obtaining accurate process labels, while the latter completely neglects the process information (e.g., action importance), leading to unfocused credit assignment. Moreover, both assign rewards at turn-level granularity, which is potentially overly fine-grained for the sub-goal spanning multiple turns \cite{guo2025segment, yin2025segmenting}. This is exemplified in Fig \ref{fig_introduction}(c-d), where sub-goal2 contains action $A_2$ and $A_3$. All of these lead to unreliable credit assignment.

In this paper, we propose (\textbf{HISR}) exploiting \textbf{H}indsight \textbf{I}nformation to modulate \textbf{S}egmental process \textbf{R}ewards for multi-turn agentic RL, which closely aligns rewards with sub-goals and underscores significant segments to enhance the reliability of credit assignment. Specifically, to avoid overly fine-grained allocation of turn-level rewards, we exploit GPT4o to split trajectories into several meaningful segments, where each segment corresponds to a sub-goal for the task. A segment-level process RM is then presented to assign rewards for each sub-goal in the task, facilitating the allocation of process rewards aligning with the sub-goal. Additionally, to emphasize significant segments in the trajectory, we devise a hindsight model that recovers each masked action in the turn to capture likelihoods, which reflects the preference of performing a certain action after knowing trajectory outcome. With this characteristic, the ratios of sequence likelihoods between the hindsight and policy model are designed to measure the action importance. The ratios are subsequently employed to aggregate segment importance scores, which in turn modulate segmental process rewards, underscoring significant segments to enhance credit assignment reliability.

Extensive experimental results on three publicly available agentic benchmarks and case studies demonstrate the effectiveness of our method. In summary, our main contributions include:

1) We propose (\textbf{HISR}) exploiting \textbf{H}indsight \textbf{I}nformation to modulate \textbf{S}egmental process \textbf{R}ewards for multi-turn RL, which closely aligns rewards with sub-goals and underscores significant segments to enhance credit assignment reliability.

2) We present a segment-level process RM to assign rewards for sub-goals in the task, avoiding overly fine-grained allocation of turn-level rewards. A hindsight model is also devised to capture likelihoods, which reflects the preference of performing a certain action after knowing trajectory outcome.

3) We design the ratios of sequence likelihoods between the hindsight and policy model to measure the action importance. The ratios are subsequently employed to aggregate segment importance scores, which in turn modulate the segmental process rewards, enhancing credit assignment reliability.

4) We carry out extensive experiments on three available agentic benchmarks. The newly achieved state-of-the-art performance and case study demonstrate the effectiveness of our method. The code will be released soon to foster future research.

\section{Related Work}

\textbf{LLM-based Agent}. It aims to equip LLMs with the capability of solving complex long-horizon agentic decision-making tasks (e.g., household assistance \cite{shridhar2020alfworld, puig2018virtualhome}, webshop browsing \cite{yao2022webshop}). To achieve this goal, some preliminary explorations have been conducted in two mainstreams: (1) prompt engineering \& supervised fine-tuning advocates adopting the data characterized by rigorous interactive paradigms to perform the behavior cloning for LLMs. For instance, \citet{yao2022react} required the model to bind reasoning and acting in an interleaved manner, stimulating their synergistic effect. \citet{shinn2023reflexion} introduced the reflection operation, alleviating the issues of hallucination and error propagation in the intermediate chain-of-thought reasoning \citep{wei2022chain}. Despite improvements, these methods fail to enable LLM to interact with environment in real time and are plagued by the acquisition of diverse and high-quality trajectory data. To overcome this issue, researchers \cite{wang2025spa, feng2025group} instead leverage (2) multi-turn agentic RL (i.e., PPO \cite{schulman2017proximal}) to optimize the decision-making of LLMs during the exploration in the environment. Compared with (1), the objective of multi-turn RL is maximizing the cumulative reward of the trajectory, which is more suitable for agentic tasks. In this paper, we adopt the latter and focus on pursuing a more reasonable reward allocation to enhance LLMs' agentic capabilities.

\noindent
\textbf{RM in Multi-turn RL}. RM plays an important role in providing supervision signals in multi-turn RL. Early research adopts the outcome-based RM \cite{ouyang2022training, schulman2017proximal}, which merely assigns a single reward at the end of trajectory according to the trajectory outcome. However, in the agentic tasks, the exacerbation of long-horizon and the characteristic of delayed reward hinder propagating the final reward to actions that occurred earlier, struggling to determine whether an action contributes positively or negatively to the trajectory outcome \cite{wang2025spa}. 

To alleviate this issue, researchers present turn-level process RM to perform credit assignment within trajectory, which can be roughly divided into two types based on the requirement of process reward labeling: (1) One employs finite MCTS \cite{guo2025segment, wang2024math, xi2025agentprm} or GPT-4 \cite{xie2025capo} to annotate pseudo process rewards. However, they face the difficulty of cost-effectively obtaining accurate process labels, yielding the noisy turn-level supervision. (2) The others are independent of process reward labels, instead leveraging \cite{kim2023preference, wang2025spa} the trajectory outcome to indirectly monitor the credit assignment. But these methods completely neglect the process information (e.g., action importance), resulting in unfocused credit assignment \cite{verma2024hindsight}. Moreover, both assign rewards at turn-level granularity, which is potentially overly fine-grained for sub-goals spanning multiple turns \cite{yin2025segmenting, guo2025segment}. All of these lead to unreliable credit assignment.

In this paper, we present segmental process rewards, which closely align the reward allocation with the sub-goal in the task, avoiding excessively granular allocation to turns. We also exploit the hindsight information to modulate rewards without extra process labels, incorporating valuable process insights to underscore significant segments and enhancing credit assignment reliability.

\section{Methodology}

\begin{figure*}[http]
    \centering
    \includegraphics[width=1\linewidth]{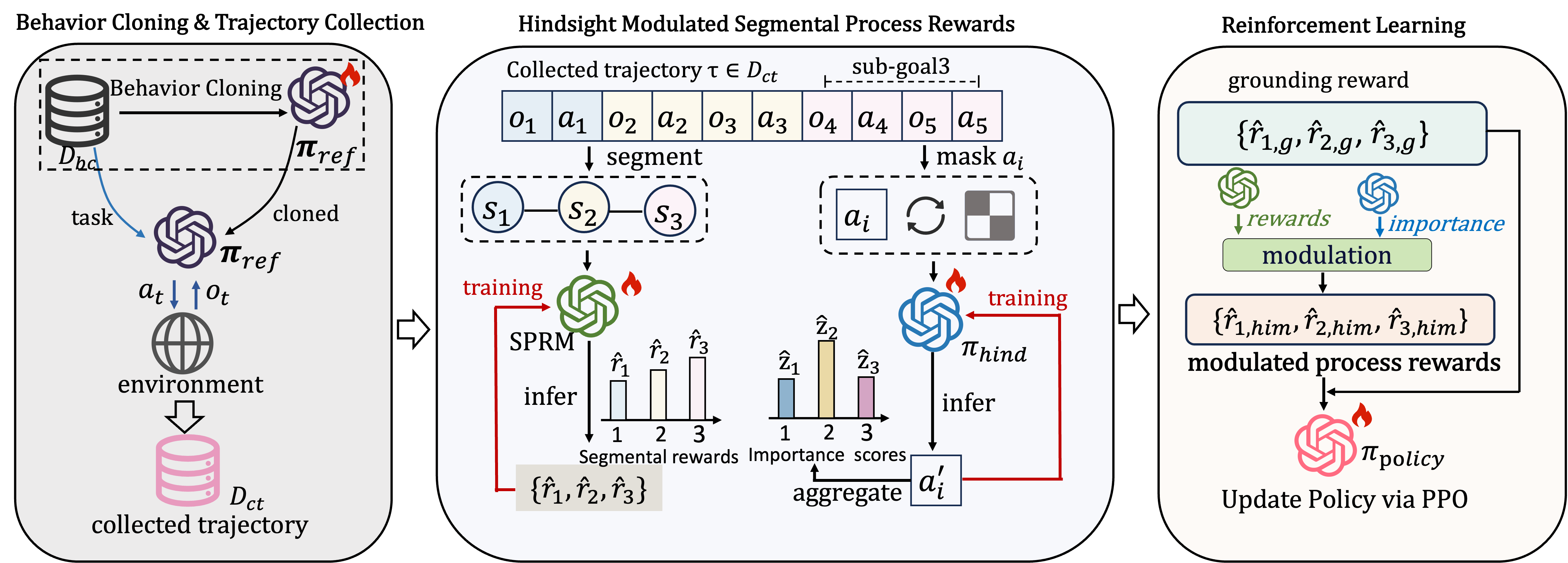}
    \caption{An illustration of our method. The left showcases the details of behavior cloning and trajectory collection. The middle illustrates the training and inference of the segment-level process reward and hindsight model. The right describes the construction of hindsight information modulated rewards and employs them for policy optimization.}
    \label{main_figure}
\end{figure*}

In this section, we elaborate on exploiting hindsight information to modulate segmental process rewards, which closely aligns rewards with sub-goals and underscores significant segments to enhance the reliability of credit assignment. As illustrated in the left of Fig \ref{main_figure}, we perform behavior cloning for LLM, equipping it with the basic task planning and reasoning capability, followed by trajectory collection. Then, as shown in the middle of Fig \ref{main_figure}, the hindsight model and segment-level process RM characterized by progress estimation are constructed based on collected trajectories. Last, we achieve segmental importance scores based on the ratios of sequence likelihoods between the hindsight and policy model, modulating segmental process rewards for multi-turn RL algorithms (i.e., PPO), as depicted at the right of Fig \ref{main_figure}.

\subsection{Behavior Cloning and Trajectory Collection}

To achieve basic capabilities of task planning and reasoning, the LLM is required to perform behavior cloning via supervised fine-tuning on the successful expert trajectories dataset $D_{bc}=\{Y_i\}_{i=1}^{|D_{bc}|}$, as shown in the left of Fig \ref{main_figure}. Particularly, let $Y_i=\{o_{i,1},a_{i,1},...,o_{i_k},a_{i,k},...o_{i_m},a_{i,m}\}$ denote the $i$-th trajectory with $m$ turns in $D_{bc}$, where $o_{i,k}$ and ${a_{i,k}}$ denote the observation returned by environment and response generated by LLM in the $k$-th turn. It should be noted that each response $a_{i, k}$ is a thought-action pair, which adheres to ReAct \cite{yao2022react} style, namely executing CoT reasoning prior to determining each action for stimulating their synergistic effect. Then, we optimize LLM by minimizing the negative log-likelihood of each trajectory $Y_i$ as below \eqref{eq1}:

\begin{small}
\begin{equation}
\label{eq1}
    \mathcal{L}_{\textup{bc}}(Y_i) = -\sum_{k=1}^{m}\sum_{j=1}^{|a_{i,k}|}log(a_{i,k}^{j}|o_{i,<=k},a_{i,<k},a_{i,k}^{<j}).
\end{equation}
\end{small}

Note that we only compute the loss of thought-action pairs $a_{i,k}$, excluding the observations $o_{i, k}$ to improve the training stability \cite{jin2025search}. By doing so, the LLM acquires the fundamental capability to accomplish the agentic tasks, resulting in a reference policy model $\pi_{ref}$, which considerably accelerates the optimization of downstream RL.

Considering that effectively evaluating the impact of a certain action on trajectory outcome requires observing this action in a wide range of interactive histories, we have the reference policy model $\pi_{ref}$ to explore in the environment, thereby collecting trajectories. Concretely, we follow the line of \citet{wang2025spa}, which conducts $N$ rollouts for each task in $D_{bc}$ without any demonstrations and manually designed exploration schemes \cite{hao-etal-2023-reasoning, zhao2023large} to ensure trajectory diversity. The samples with failed instruction following or repetition are further filtered to achieve the collected trajectory dataset $D_{ct}=\{\{({\tau}_{i,j},R_{i,j})\}_{j=1}^{N}\}_{i=1}^{|D_{ct}|}$, where ${\tau}_{i,j}$ and $R_{i,j}$ denote the $j$-th trajectory for the $i$-th task and its corresponding trajectory outcome (i.e., a scalar number to reflect the task completion).

\subsection{Hindsight Information Modulated Segmental Process Rewards}

To prevent potentially overly fine-grained process rewards caused by
mere reward allocation at turn-level granularity and unfocused process rewards caused by the complete neglect of process information, we propose hindsight information modulated segmental process rewards centered on aligning rewards with sub-goals and underscoring significant segments to enhance credit assignment reliability, as shown in the middle and right of Fig \ref{main_figure}. 

\noindent
\textbf{Segmental Process Reward Model (SPRM).} The SPRM is presented to perform credit assignment within the trajectory at the segment-level granularity. Here, a segment refers to a turn or a contiguous set of turns in the trajectory, typically corresponding to a sub-goal in the decomposition of the given task. To obtain segments contained in each trajectory $\tau$ in $D_{ct}$, we exploit GPT-4o \cite{hurst2024gpt} to identify sub-goals in the task, and allocate turns to the corresponding sub-goal, achieving the segmented trajectory dataset $D_{st}=\{\{({\tau}_{i,j}^{s},R_{i,j})\}_{j=1}^{N}\}_{i=1}^{|D_{st}|}$, where $\tau^{s} = \{s_1, s_2, ...,s_n\}$ denote the trajectory with $n$ segments. With $D_{st}$, we borrow the idea from progress estimation \cite{wang2025spa} and further perform credit assignment at the segment-level granularity. For each segmented trajectory $\tau^{s}$, SPRM relies on the reasoning capability of the reference policy model $\pi_{ref}$ and intuitively decomposes the scalar trajectory outcome $R$ into segmental contributions. It reflects how each segment incrementally propels the agent towards trajectory outcome. Concretely, a lightweight Multi-Layer Perceptron (MLP) is appended to the last hidden layer of $\pi_{ref}$, allowing it to output a scalar contribution score $r_i$ at the end of each segment as \eqref{eq2}. Furthermore, for a trajectory $\tau^{s}$ with $n$ segments, we optimize the SPRM by minimizing the mean square error loss as follows \eqref{eq3}: 

\begin{small}
\begin{align}
    & \label{eq2} r_i = W_2(\textup{SiLU}(W_1h_i)), \\
    & \label{eq3} \mathcal{L}_{\textup{sprm}}(\tau^{s}) = (R-\sum_{i=1}^{n}r_i)^2,
\end{align}
\end{small}

\noindent
where $h_i$, $\textup{SiLU}$, $W_1$, $W_2$, denote the hidden state of the last token in $i$-th segment $s_i$, the activation function and learnable weights in MLP.

\noindent
\textbf{Hindsight Information Modulation.} With SPRM, we realize credit assignment at the segment-level granularity, which closely aligns rewards with sub-goals in the task. To further conquer the issue of unfocused process rewards caused by the complete neglect of process information and effectively incorporate valuable process insights, we introduce hindsight information, which refers to the importance of a certain action towards achieving the trajectory outcome in hindsight. We leverage the hindsight information to modulate segmental process rewards, making the credit assignment underscoring the significant segments within the trajectory. 

To model the hindsight information, the reference policy model $\pi_{ref}$ undergoes continued training with an objective akin to masked language modeling, achieving a hindsight model $\pi_{hind}$. Particularly, for a collected trajectory $\tau =\{o_{1},a_{1},...,o_{k},a_{k},...o_{m},a_{m}\}$ with $m$ turns, we mask each response $a_{k}$ in the turn with an underline and demand $\pi_{ref}$ to recover it as \eqref{eq4}, capturing likelihoods which reflects the preference of performing a certain action after knowing the outcome of the trajectory.  

\begin{small}
\begin{equation}
\label{eq4}
    \mathcal{L}_{\textup{hind}}(a_{k}) = -\sum_{j=1}^{|a_{k}|}log(a_{k}^{j}|o,a_{<k},a_{>k},a_{k}^{<j}),
\end{equation}
\end{small}

\noindent
with this characteristic, we draw inspiration from \citet{harutyunyan2019hindsight} and design the ratios of sequence likelihoods between the hindsight $\pi_{hind}$ and policy model $\pi_{policy}$ to measure the importance $z(a_{k})$ of $k$-th action, as formalized in \eqref{eq5}, \eqref{eq6}.

\begin{small}
\begin{align}
    & \label{eq5} r(a_{k}^j)=\frac{\pi_{hind}(a_{k}^j|o,a_{<k},a_{>k},a_{k}^{<j})}{\pi_{policy}(a_{k}^j|o_{<=k},a_{<k},a_{k}^{<j})}, \\
    & \label{eq6} z(a_{k}) = exp(\frac{1}{\beta|a_{k}|}\sum_{j=1}^{|a_{k}|}log(r(a_{k}^j)),
\end{align}
\end{small}

\noindent
where $a_{k}^j$ denote $j$-th token in the $k$-th action. $\beta$ is the hyper-parameter to control the distinction between different actions. Note that $\pi_{policy}$ computes likelihoods of the current action $a_{k}$ with only information available prior to $a_{k}$, while $\pi_{hind}$ additionally incorporates the subsequent information available in hindsight. Intuitively, if $z(a_{k})>1$, it means the agent prefers to execute the current action $a_{k}$ in hindsight, indicating that this action propels the agent towards the trajectory outcome, highlighting its importance and vice versa. Consequently, for a trajectory $\tau$ with $m$ turns, we can obtain $m$ turn-level action importance ratios $z=\{z(a_1),z(a_2),...,z(a_m)\}$. Similarly, to align importance measurement with the sub-goal in the task, we aggregate segmental importance scores $z_s=\{z(s_1),z(s_2),...z(s_n)\}$ by adding up turn ratios belonging to the same segment, highlighting the significant segments in the trajectory.

To coordinate segmental process rewards and importance scores, for a trajectory $\tau$ with $n$ segments, we leverage trained SPRM to predict segmental progress rewards $\hat{R}=\{\hat{r}_1, \hat{r}_2, ..., \hat{r}_n\}$. Then $\pi_{hind}$ and $\pi_{policy}$ are exploited to compute segmental importance scores $\hat{z}_s=\{\hat{z}_{s_1}, \hat{z}_{s_2},..., \hat{z}_{s_n}\}$. We perform multiplication and normalization on $\hat{R}$ and $\hat{z}_s$ to scale rewards based on importance scores,  achieving hindsight information modulated segmental process rewards $\hat{R}_{him}=\{\hat{r}_{1}^{him}, \hat{r}_{2}^{him},...,\hat{r}_{n}^{him}\}$. By doing so, the valuable process insights (i.e., action importance) are incorporated, and significant segments would be underscored, improving the credit assignment reliability.

\begin{small}
\begin{equation}
\label{eq7}
    \hat{R}_{him} = \frac{\hat{R} \odot \hat{z}_s}{||\hat{R} \odot \hat{z}_s||}.
\end{equation}
\end{small}

\subsection{Multi-turn Agentic RL}

Modulated by hindsight information, $\hat{R}_{him}$ accurately reflects how each segment propels the agent towards the trajectory outcome. But it fails to indicate the executability of each action in the environment. To bridge this gap, we additionally introduce an action grounding reward $\hat{r}^g$ that takes 1 if the action is executable, else 0, achieving $\hat{r}^{fuse}$:

\begin{small}
\begin{equation}
\label{eq8}
    \hat{r}^{fuse} = (1-\alpha)\hat{r}^{him} + \alpha \hat{r}^g,
\end{equation}
\end{small}

\noindent
where $\alpha$ is the hyper-parameter to balance two rewards. Later, $\hat{r}_{fuse}$ is used to guide the optimization of policy model $\pi_{policy}$ under PPO algorithm:

\begin{small}
\begin{align}
    \mathcal{L}_{\textup{clip}}(\theta) &= \mathbb E_{t} \bigg[ \min ( \frac{\pi_\theta(a_t \mid s_t)}{\pi_{\theta_{old}}(a_t \mid s_t)} \hat{A}_{t}^{fuse}, \nonumber
    \\& \operatorname{clip}( \frac{\pi_\theta(a_t \mid s_t)} {\pi_{\theta_{old}}(a_t \mid s_t)}, 1-\epsilon, 1+\epsilon) \hat{A}_{t}^{fuse} \bigg],
\end{align}
\end{small}

\noindent
where $\pi_{\theta}$ is the policy with parameters $\theta$. $\hat{A}_{t}^{fuse}$ is the advantage, which is adopted as generalized advantage estimation (GAE). It blends temporal-difference errors $\delta_t=\hat{r}_{t}^{fuse}+\gamma V_{\phi}(s_{t+1})-V_{\phi}(s_t)$. $V_{\phi}$ is an alternatively trained value network.

\section{Experiments}

In this section, we introduce datasets, baselines, and experimental settings. Then, we present experimental results and provide a detailed analysis.

\subsection{Datasets and Experimental Settings}

To verify the effectiveness of our method, we carry out experiments on three publickly available agentic benchmarks: Alfworld \cite{shridhar2020alfworld}, Virtualhome \cite{puig2018virtualhome} for embodied household suits, and Webshop \cite{yao2022webshop} for web navigation. Particularly, for Alfworld, we follow previous works \cite{wang2025spa, wang-etal-2025-steca, feng2025group} to adopt the version constructed by \citet{song-etal-2024-trial}. For Virtualhome, we employ the filtered version provided by \citet{wang-etal-2025-steca}. On all benchmarks, the agent receives an environmental observation each turn and accordingly determines an action until completing the task or until reaching maximum interactive turns, ultimately achieving a scalar score as the trajectory outcome. Further details about datasets and specific experimental settings can be found in Appendix \ref{appendix.a} and \ref{appendix.b}. 

\begin{table*}[htbp]
\centering
\resizebox{1.0\textwidth}{!}{
\begin{tabular}{lccccccccc@{\hskip 1.5em}cc}  
\toprule
\multirow{2}{*}{Methods} & \multirow{2}{*}{Type} & \multicolumn{7}{c}{Alfworld $\uparrow$} & \multirow{2}{*}{Virtualhome $\uparrow$} & \multirow{2}{*}{Webshop $\uparrow$}\\
\cmidrule{3-9}
&  & PICK $\uparrow$ & CLEAN $\uparrow$ & HEAT $\uparrow$ & COOL $\uparrow$ & LOOK $\uparrow$ & PICK2 $\uparrow$ & \textit{\textbf{Avg}} $\uparrow$ &  &  \\
\midrule
Llama3.2 \cite{dubey2024llama} & PE & 12.5 & 0.0 & 0.0 & 0.0 & 0.0 & 0.0 & 2.3 & 1.2 & 19.8 \\
GPT4o \cite{hurst2024gpt} & PE & 75.3 & 60.8 & 31.2 & 56.7 & 21.6 & 49.8 & \textit{48.0} & 20.8 & 23.7 \\
Gemini2.5pro \cite{team2023gemini} & PE & 92.8 & 63.3 & 62.1 & 69.0 & 26.6 & 58.7 & \textit{60.3} & 31.7 & 35.9 \\
\midrule
SFT \cite{ouyang2022training} & BC & 79.2 & 77.4 & 73.9 & 61.9 & 83.3 & 58.8 & \textit{73.1} & 51.8 & 62.0 \\
RFT \cite{tang2024mathscale} & BC & 83.3 & 83.9 & 78.3 & 66.7 & 77.8 & 47.1 & \textit{74.6} & 53.1 & 62.9 \\
DPO \cite{rafailov2023direct} & BC & 87.5 & 83.9 & 60.9 & 66.7 & 88.9 & 64.7 & \textit{76.1} & 52.8 & 62.6 \\
\midrule
PPO \cite{schulman2017proximal} & RL & 83.3 & \textbf{87.1} & 73.9 & 61.9 & 77.8 & 47.1 & \textit{73.9} & 51.0 & 62.1 \\
GRPO \cite{shao2024deepseekmath} & RL & 87.5 & 70.9 & \underline{82.6} & 57.1 & \underline{88.9} & 52.9 & \textit{73.4} & 51.2 & 61.8 \\
Archer \cite{zhou2024archer} & RL & 83.3 & \underline{83.9} & 65.2 & \textbf{85.7} & 66.7 & \underline{58.8} & \textit{75.4} & -- & -- \\
StepAgent \cite{deng2024novice} & RL & 83.3 & \textbf{87.1} & 78.3 & \underline{71.4} & 77.8 & 41.2 & \textit{75.4} & -- & -- \\
RAGEN \cite{wang2025ragen} & RL & \underline{91.7} & 77.4 & 78.3 & 57.1 & \underline{88.9} & 52.9 & \textit{75.4} & 52.1 & 63.0 \\
PRM4A \cite{choudhury2025process} & RL & 58.3 & 80.6 & 73.9 & \underline{71.4} & \textbf{100.0} & \underline{58.8} & \textit{73.9} & -- & -- \\
SPA \cite{wang2025spa} & RL & \textbf{95.8} & \underline{83.9} & \textbf{87.0} & 61.9 & 77.8 & \underline{58.8} & \textit{\underline{79.1}} & \underline{53.4} & \underline{64.1} \\
\midrule
HISR & RL & 83.3 & \textbf{87.1} & 65.2 & \textbf{85.7} & \textbf{100.0} & \textbf{82.4} & \textbf{\textit{83.6}} & \textbf{59.1} & \textbf{69.1} \\
-w/o HIM & RL & 87.5 & 83.9 & 78.3 & 66.7 & 94.4 & 70.6 & \textit{80.6} & 55.1 & 63.7 \\ 
-w/o SPR & RL & 87.5 & 83.9 & 78.3 & 76.2 & 94.4 & 70.6 & \textit{82.1} & 57.9 & 69.1 \\ 
-w/o BOTH & RL & 87.5 & 83.9 & 69.6 & 71.4 & 94.4 & 58.8 & \textit{78.4} & 53.4 & 63.7 \\-w/o AGS & RL & 91.7 & 87.1 & 78.3 & 66.7 & 100.0 & 64.7 & \textit{82.8} & 57.5 & 68.6 \\
\bottomrule
\end{tabular}
}
\caption{Evaluation results and ablation study on three benchmarks. \textit{Avg} indicates the comprehensive score in the Alfworld dataset. PE, BC, and RL indicate methods of prompt engineering, behavior cloning, and reinforcement learning. Best in bold, the runner-up with an underline. The top three parts denote baseline methods, while the last part denotes our method and its ablation variants.  
$*$ denotes it adopts turn-level process rewards.
}  
\label{main_experiment}
\end{table*}

\subsection{Baselines}

We compare our method with the following three types of baselines. (1) Prompt engineering evaluates the frozen LLMs (\textbf{Llama3.2} \cite{dubey2024llama}, \textbf{GPT4o} \cite{hurst2024gpt}, and \textbf{Gemeni2.5pro} \cite{team2023gemini}) under zero-shot setting. (2) Behavior cloning includes the methods of supervised fine-tuning (\textbf{SFT} \cite{ouyang2022training}), reject sampling fine-tuning (\textbf{RFT} \cite{tang2024mathscale}), and direct preference optimization (\textbf{DPO} \cite{rafailov2023direct}). We serve them as baselines to evaluate the performance gain from offline samples. (3) Reinforcement Learning based Fine-tuning is our primary baseline. The methods of \textbf{PPO} \cite{schulman2017proximal}, \textbf{GRPO} \cite{shao2024deepseekmath}, \textbf{Archer} \cite{zhou2024archer} with outcome-based rewards are selected to compare with the following methods characterized by process rewards. \textbf{StepAgent} \cite{deng2024novice} leverages the expert policy to score each turn, achieving the turn-level supervision. \textbf{RAGEN} \cite{wang2025ragen} designs a bi-level GAE, providing fine-grained rewards in different turns. \textbf{PRM4A} \cite{choudhury2025process} and \textbf{SPA} \cite{wang2025spa} are the strongest baselines with turn-level process rewards, which leverage trajectory outcome to indirectly monitor the credit assignment. More details can be found in Appendix \ref{appendix.c}.

\subsection{Experimental Results}

The experimental results on three benchmarks are reported in Table \ref{main_experiment}. We can observe that: (1) Even the best-performing model (\textbf{Gemini2.5pro}) exhibits far from satisfactory with the method of prompt engineering, revealing the deficiency of LLM in solving complex long-horizon decision-making agentic tasks. (2) When fine-tuned on the expert trajectory dataset $D_{bc}$, the performance (\textbf{SFT}) substantially surges across benchmarks, demonstrating the effectiveness of equipping LLM with basic task planning and reasoning capabilities. However, further employing more high-quality trajectories in \textbf{RFT} or pairwise preference data in \textbf{DPO} can not break the performance bottleneck. We believe this is because the offline data is insufficient to represent the complex decision-making of the LLM-based agent in the environment.  

(3) Compared to \textbf{SFT}, the RL methods of \textbf{PPO} and \textbf{GRPO} with outcome-based reward only exhibit a slight improvement, even suffering from degradation in some cases (e.g., Virtualhome dataset). This can be ascribed to the characteristic of long-horizon in the agentic task and sparse delayed reward in the algorithm, which makes it difficult to propagate the final reward to the actions that happened in the early turns. (4) When equipped with dense process rewards, this deficiency is partially alleviated, and performance gains are obtained (e.g., \textbf{Archer}, \textbf{StepAgent}, \textbf{RAGEN}). This is consistent with the observation in \citet{wang2025ragen}, where fine-grained reward plays an important role in multi-turn agentic RL. (5) Compared to the strongest baseline \textbf{SPA}, our method outperforms on all benchmarks. This can be attributed to two factors. One is closely aligning credit assignment with sub-goals in the task, avoiding excessively granular allocation to turns. Another is incorporating valuable process insights (i.e., action importance) to prevent unfocused rewards, where we introduce hindsight information to modulate segmental process rewards, underscoring significant segments within the trajectory. In the end, the considerable performance gains across benchmarks demonstrate the validity of our method.

\begin{table*}[t]
\centering
\scriptsize
\setlength{\tabcolsep}{2pt}
\renewcommand{\arraystretch}{1.2}

\definecolor{rowgray}{gray}{0.95}

\renewcommand{\tabularxcolumn}[1]{m{#1}}

\newcolumntype{Y}{>{\raggedright\arraybackslash}X}
\newcolumntype{C}{>{\centering\arraybackslash}c}
\resizebox{0.95\textwidth}{!}{
\begin{tabularx}{\textwidth}{@{} C @{\hspace{1em}} Y @{\hspace{1em}} C @{}}
\toprule
\textbf{Segment (turn)} &
\multicolumn{1}{c}{\textbf{Conversations}} &
$\mathbf{(\hat{R},\hat{z}, \hat{R}_{him})}$ \\
\midrule

\multicolumn{3}{@{}c}{\textbf{Case: Clean and put bowl $\rightarrow$ Fridge}} \\
\midrule

\rowcolor{rowgray}
\makecell[c]{\textbf{Segment1 (turn1)}\\[2pt] Task interpretation \\ and decomposition} &
\textbf{Task Define}: You are in the middle of a room... \textbf{Your task is to:} clean some bowl and put it in fridge. \par \vspace{2pt}
\textbf{Thought}: The task is to clean a bowl and put it in the fridge. I should first locate the fridge to find a bowl that needs cleaning. \textbf{Action}: go to fridge 1.
& $(0.069,\ 0.127,\ 0.039)$ \\

\multirow{2}{*}[-1em]{\makecell[c]{\textbf{Segment2 (turn2-3)}\\ Opening the fridge \\ and picking up \\ the bowl}} &
\textbf{Observation}: The fridge 1 is closed. \par \vspace{2pt}
\textcolor{blue}{\textbf{Thought}: I'm in front of the fridge, but it's closed. I need to open it to find a bowl. \textbf{Action}: open fridge 1.}
& \multirow{2}{*}[-2em]{\makecell[c]{$(\textcolor{red}{0.118},\ \textbf{0.392},\ \textcolor{red}{\textbf{0.205}\uparrow})$}} \\

&
\textbf{Observation}: You open the fridge 1. The fridge 1 is open. In it, you see a apple 3, a apple 2, a apple 1, a bowl 1, a bread 1, and a cup 1. \par \vspace{2pt}
\textcolor{blue}{\textbf{Thought}: The fridge is now open, and I see a bowl inside. I should take the bowl out to clean it. \textbf{Action}: take bowl 1 from fridge 1.}
& \\

\cellcolor{rowgray} & 
\cellcolor{rowgray}
\textbf{Observation}: You pick up the bowl 1 from the fridge 1. \par \vspace{2pt}
\textbf{Thought}: With the bowl in hand, I need to find a place to clean it. The sink is the best place for this task. \textbf{Action}: go to sinkbasin 1.
& \cellcolor{rowgray} \\

\cellcolor{rowgray}\multirow{-2}{*}[2em]{\makecell[c]{\textbf{Segment3 (turn4-5)}\\ Going to the sink \\ and cleaning \\ the bowl}} &
\cellcolor{rowgray}
\textbf{Observation}: On the sinkbasin 1, you see nothing. \par \vspace{2pt}
\textbf{Thought}: I'm at the sink and it's clear, which is perfect for cleaning the bowl without any obstructions. \textbf{Action}: clean bowl 1 with sinkbasin 1.
& \cellcolor{rowgray}\multirow{-2}{*}[1.0em]{\makecell[c]{$(0.132,\ 0.286,\ 0.167)$}} \\ 

\multirow{2}{*}[-0.5em]{\makecell[c]{\textbf{Segment4 (turn6-7)}\\ Returning to \\ the fridge and \\ placing the bowl}} &
\textbf{Observation}: You clean the bowl 1 using the sinkbasin 1. \par \vspace{2pt}
\textbf{Thought}: The bowl is now clean, and I need to put it back in the fridge as per the task instructions. I should head back to the fridge. \textbf{Action}: go to fridge 1.
& \multirow{2}{*}[-1.5em]{\makecell[c]{$(0.681,\ 0.195,\ 0.589)$}} \\

&
\textbf{Observation}: The fridge 1 is open. In it, you see a apple 3, a apple 2, a apple 1, a bread 1, and a cup 1. \par \vspace{2pt}
\textbf{Thought}: I'm back at the fridge, which is still open from earlier. I can now place the clean bowl inside. \textbf{Action}: put bowl 1 in/on fridge 1.
& \\

\bottomrule
\end{tabularx}}

\caption{Case study on the Alfworld dataset. The left shows segmented results of the trajectory, where Segment (turn) denotes consecutive turns contained in the segment. The middle shows the explored trajectory, where blue texts highlight the highest importance scores segment. The right shows the tuple of raw segmental process rewards, hindsight information modulated rewards, and segmental importance ratio scores.}
\label{case_study_main}
\end{table*}

\subsection{Ablation Study}

To explicitly illustrate the efficacy of our method, we conduct ablation studies to validate its core design on all benchmarks. Specifically, as shown in the last part of Table \ref{main_experiment}, we define the following ablation variants: \textbf{(1) -w/o HIM} removes hindsight information modulation, retaining only segmental process rewards to align the credit assignment with sub-goals in the task. \textbf{(2) -w/o SPR} removes segmental process rewards, retaining only the hindsight information modulation to underscore the significant turns in the trajectory. \textbf{(3) -w/o BOTH} is the combination of \textbf{(1)} and \textbf{(2)}. \textbf{(4) -w/o AGR} removes the action grounding reward, keeping other core designs unchanged. Note that our method is equal to the variant -w/o SPR in the Webshop dataset. This is because GPT-4o deems that $89.3\%$ samples do not require segmentation as shown in Appendix \ref{appendix.b}. It demonstrates that most turns have already been aligned with the sub-goals in the task. Thus, we adopt the turn-level process reward. 

Concretely, we can draw inferences according to the results of Table \ref{main_experiment}. (1) Removing hindsight information modulation (-w/o HIM) results in a performance drop, reflecting its crucial role in incorporating valuable insights to underscore significant segments in the trajectory. (2) Replacing segmental process rewards also leads to a performance degradation, underscoring its vital role in closely aligning rewards with sub-goals to prevent excessively granular allocation to turns. 
(3) Removing both hindsight information and segmental process rewards (-w/o BOTH) leads to the worst performance, indicating the necessity of each core design. (4) Removing action grounding reward (-w/o AGS) impairs performance, demonstrating the validity of making rewards represent action executability. 


\begin{figure}[h]
    \centering
    \includegraphics[width=1\linewidth]{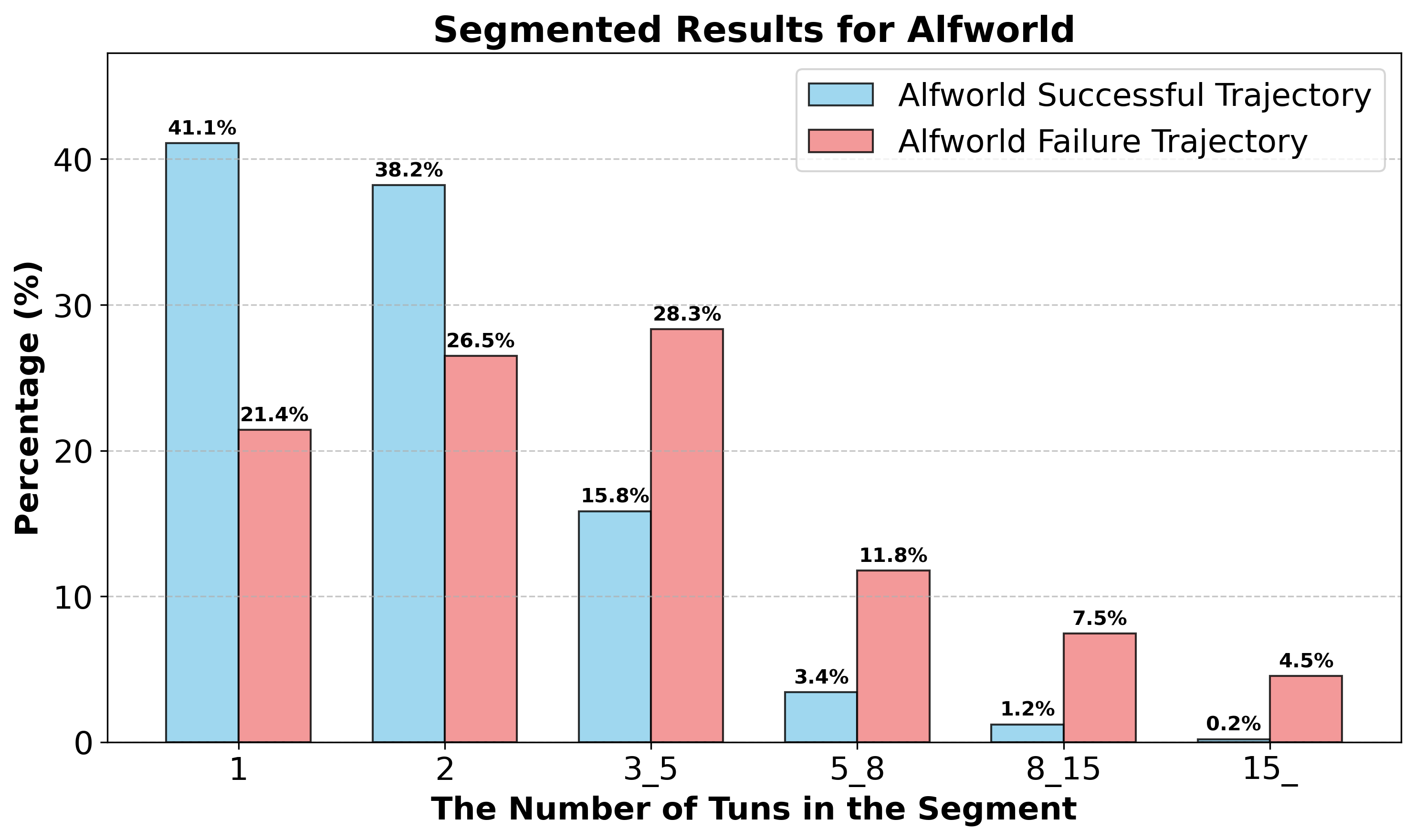}
    \caption{Statistical results of the number of turns in the segment of the Alfworld dataset.}
    \label{segmented_result}
\end{figure}

\begin{figure}[h]
    \centering
    \includegraphics[width=1\linewidth]{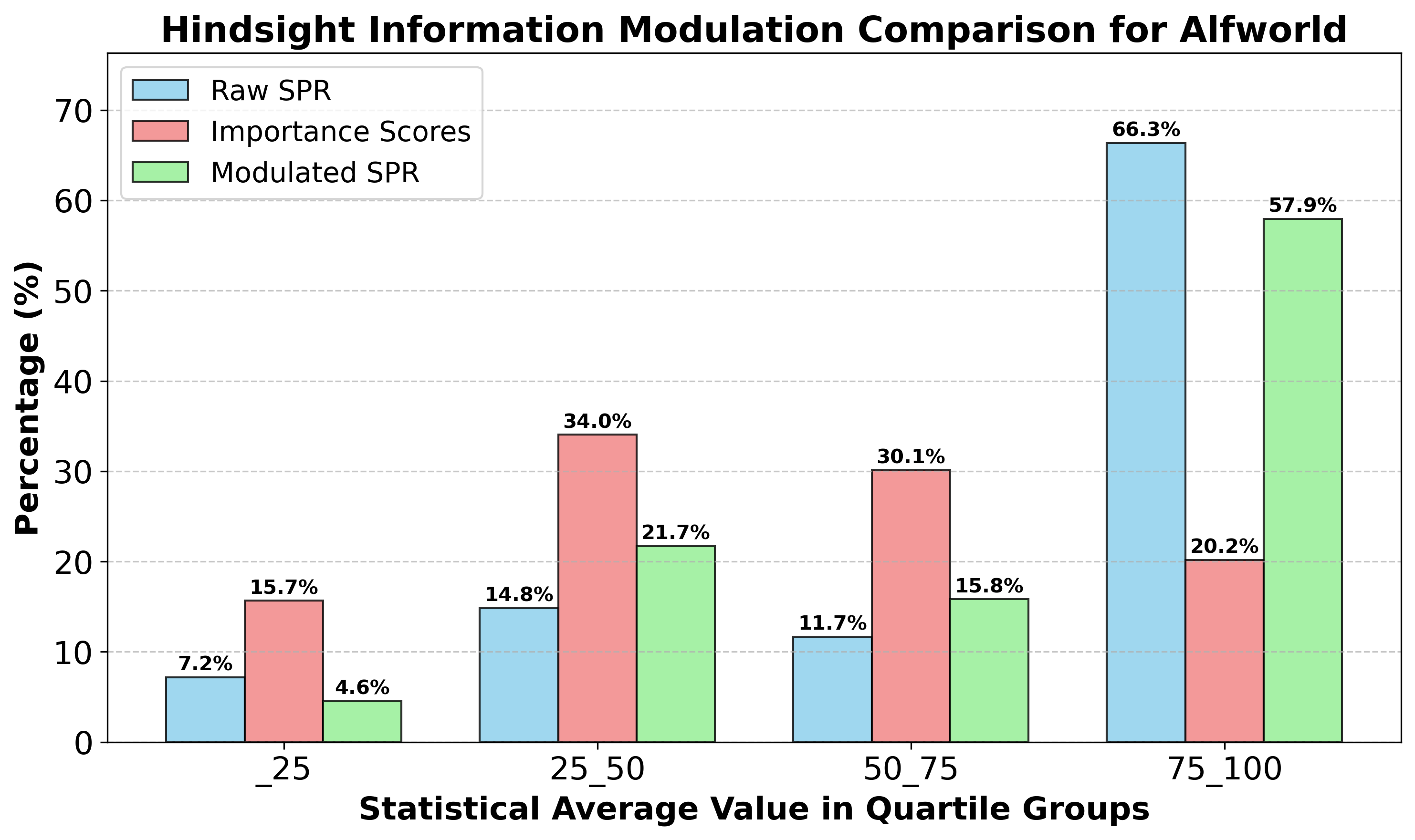}
    \caption{Statistical results of the process rewards distribution in the Alfworld datasets. SPR denote the segmental process rewards.}
    \label{hindsight_information_result}
\end{figure}

\subsection{Analysis of SPR and HIM}

To intuitively understand each core design, we conduct two statistical analyses on the Alfworld dataset. Specifically, as shown in Fig \ref{segmented_result}, we compute the number of turns in all segments and observe that GPT-4o merged most of the turns in the trajectories (segments containing two or more turns accounted for over 50\%), which demonstrates that the phenomenon of a sub-goal spanning on multiple turns is common, reflecting the feasibility of segmental process rewards. This phenomenon is more highlighted in the failure trajectory. We further scrutinize the trajectory and find that GPT-4o combines multiple trial-and-error turns aimed at achieving the same sub-goal into a single segment. Additionally, we can observe in Fig \ref{hindsight_information_result} that segment-level process RM inclines to allocate more rewards to the latter segment (i.e., 66.3\%). This is probably because later segments are closer to the task completion without consideration of process information. However, this prior underestimates the importance of earlier segments towards propelling the trajectory outcome, resulting in unfocused credit assignment. Instead, the important scores modulate SPR, facilitating underscoring earlier significant segments (i.e., 14.8\% to 21.7\% and 11.7\% to 15.8\%) in the trajectory. 

\subsection{Case Study}
We conduct the case study on the Alfworld dataset to illustrate our core designs, as shown in Table \ref{case_study_main}. Concretely, we observe that GPT-4o tends to merge consecutive turns so that combined segments have more complete and clear sub-goals. For instance, in segment2, the goal of opening the fridge in turn2 is picking up the bowl in turn3, which is planned in segment1 \textit{\textbf{"I should first locate the fridge to find a bowl ..."}} in advance. Compared to turn-level granularity, the semantics within each segment are more coherent, and the semantic distinctions between adjacent segments are more pronounced (corresponding to different sub-goals), which facilitates the reward model identifying the contribution of each segment towards the trajectory outcome. 
Moreover, turn-level rewards underestimate the crucial role of picking up the bowl in segment2 and cleaning the bowl in segment3 for completing the task, resulting in unfocused process rewards. Instead, our method leverages the importance scores $\hat{z}$ (i.e., $0.392$ and $0.286$) to modulate the segmental process rewards $\hat{R}$, underscoring significant segments (e.g., $0.118$ to $0.205$ and $0.132$ to $0.167$) and achieving more reasonable rewards $\hat{R}_{him}$ for RL training. More cases are shown at Table \ref{case_study_appendix_1}-\ref{case_study_appendix_2} in Appendix.

\section{Conclusions}
In this paper, we focus on improving LLMs' agentic capability by overcoming the issues of delayed propagation in sparse outcome rewards and unreliable credit assignment with potentially overly fine-grained and unfocused turn-level process rewards. Specifically, we propose (\textbf{HISR}) exploiting \textbf{H}indsight \textbf{I}nformation to modulate \textbf{S}egmental process \textbf{R}ewards for multi-turn RL, which closely aligns rewards with sub-goals and underscores significant segments to enhance the reliability of credit assignment. Extensive experimental results on three available agentic benchmarks and case study demonstrate the effectiveness of our method.

\section*{Limitations}

Despite the impressive results of our method, we have to admit our work has the following limitations: (1) to align reward allocation with sub-goals in the task, we employ GPT-4o to split the trajectory into several segments, where each segment corresponds to a sub-goal for the task. Note that we do not utilize the internal knowledge of GPT-4o to allocate rewards and improve performance, merely segmenting the trajectory. However, it still introduces an additional step in the pipeline of RL. To overcome this issue, we plan to explore automatic segmentation of trajectory based on the change in entropy during the process of rollout. (2) to address the issue of unfocused credit assignment caused by the complete neglect of process information and effectively incorporate valuable process insights, we introduce the hindsight information (i.e., segment importance scores) to underscore significant segments and achieve more reasonable segmental process rewards. However, the hindsight model is trained in advance and keeps frozen in the subsequent stage of RL training, which may lead to limited adaptation to changes in the data distribution, impairing the performance. We believe training hindsight model online can further bootstrap performance. (3) Although the importance scores achieved by the hindsight model underscore the significant segments, the prior on raw segmental process rewards is too strong (e.g., assigning high rewards to the latter segments). This causes the modulation effect from hindsight information to not be utilized to its fullest potential.
In the future, we will explore the online hindsight model, mining more hindsight information, and directly employ the hindsight information to guide the training of process reward model.



\bibliography{custom}

\appendix

\section{Details on Datasets}
\label{appendix.a}
\textbf{Alfworld} \cite{shridhar2020alfworld} is an embodied environment designed to assess the agentic capability of LLMs to solve complex long-horizon decision-making tasks. It provides interactive TextWorld environments, which are meticulously aligned with the ALFRED \cite{shridhar2020alfworld}. In each episode, the agent receives the text goal and is required to accomplish it through multi-turn interaction with the environment until completing the task or until reaching the maximum interactive turns, ultimately achieving a scalar score as the outcome of the trajectory. Here, the scalar score of '1' and '0' denote success and failure, respectively. Alfworld includes six categories of common household activities: Pick \& Place (Pick), Examine in Light (Look), Clean \& Place (Clean), Heat \& Place (Heat), Cool \& Place (Cool), and Pick Two \& Place (Pick2). In this work, we adopt the version constructed by \citet{song-etal-2024-trial} and set the maximum interactive turns as 40.

\noindent
\textbf{Virtualhome} \cite{puig2018virtualhome} is also an embodied environment designed to assess the agentic capability of LLMs akin to the Alfworld dataset. It encompasses 292 high-level household tasks and 1,374 unique action plans across 6,201 diverse virtual environments. For each task, the agent receives a high-level task, a descriptive explanation, and a set of executable actions. Then, the agent continuously selects an executable action and receives feedback from the environment until the task is completed or the maximum number of interaction turns is exceeded, ultimately taking '1' if the task is succeeded, else '0'. In this work, we adopt the filtered version provided by \citet{wang2025spa} and set the maximum interactive turns as 40.

\noindent
\textbf{Webshop} \cite{yao2022webshop} is a complex, web-based interactive environment designed to test the LLM agents in realistic online shopping scenarios \cite{guo2025segment}. In the environment, the agent is required to search for, navigate to, and ultimately execute a "buy" action when a purchase decision is made. Then, the agent receives a float number between '0' and '1', which is determined by the alignment of the selected product's attributes and price within the given instructions. In this work, we set the maximum interactive turn as 10.

To the end, we ensure that our dataset setting is consistent with the strongest baseline \textbf{SPA} \cite{wang2025spa} for a fair comparison. The specific statistical results are shown in the Table \ref{benchmark_statistical_result} as below. Note that the webshop dataset does not have fixed available actions. However, it can only choose the action types of search and click.

\begin{table}[htbp]
\centering
\resizebox{0.5\textwidth}{!}{
\begin{tabular}{lcccc}
\toprule
\textbf{Benchmark} & \textbf{Train} & \textbf{Test} & \textbf{Available Actions} & \textbf{Maximum Turns} \\
\midrule
ALFWorld & 2,851 & 134 & 13 & 40 \\
VirtualHome & 4,920 & 247 & 40 & 40 \\
Webshop & 1938 & 200 & - & 10 \\
\bottomrule
\end{tabular}
}
\caption{The statistical results of three agentic benchmarks. Train and Test denote the number of expert trajectory sample and test sample, respectively.}
\label{benchmark_statistical_result}
\end{table}

\section{Experimental Settings}
\label{appendix.b}
To make a fair comparison with other baselines, we follow them to select Llama3.2-3B-Instruct \cite{dubey2024llama} as the backbone. During behavior cloning, the backbone is trained for three epochs with the learning rate and batchsize of $(1.5e-5, 32), (4e-5, 24), (1.5e-5, 32)$ for Alfworld, Virtualhome, and Webshop datasets. Consuqentaly, we obtain the reference policy model $\pi_{ref}$ \eqref{eq1}. With $\pi_{ref}$, we conduct $N=10$ rollouts for each task to obtain the collected trajectory dataset $D_{ct}$ with a temperature of $0.7$. Then, we filter trajectories with wrong format and repetition, obtaining 26,433, 49,199, and 14,344 samples for Alfworld, Virtualhome, and Webshop datasets.

With $D_{ct}$, we exploit GPT4o to split each trajectory into a set of segments, achieving $D_{st}$. During segmentration, GPT4o believes that $89.3\%$ of the trajectories in Webshop do no need to segment, while Alfworld and Virtualhome are $11.5\%$ and $16.9\%$. This is because the interactive turns of Webshop are shortest so that most turns have already been aligned with the sub-goals in the task. Then, we continue training the $\pi_{policy}$ with $D_{st}$ for 1 epoch with the learning rate and batchsize of $(2e-6, 16), (2e-6, 16), (2e-6, 48)$ for Alfworld, Virtualhome, and Webshop datasets \eqref{eq3}. Later, we obtain the segmental process reward model. We also continue training the $\pi_{policy}$ with $D_{st}$ for 1 epoch with the learning rate and batchsize of $(1e-6, 192), (1e-6, 192), (1e-6, 24)$ for Alfworld, Virtualhome, and Webshop datasets \eqref{eq4}. The hindsight model $\pi_{hind}$ is then achieved. The $\beta$ and $\alpha$ are intuitively set to 0.3 when inferring the hindsight information \eqref{eq6} and introducing the action grounding reward \eqref{eq8}. During RL, we adopt the Low-Rank Adaptation (Lora \cite{DBLP:conf/iclr/HuSWALWWC22}) with $rank=8$. The coefficient of kl divergence, $\gamma$, and $\lambda$ in PPO algorithm are set $1e-2$, $0.99$, and $0.95$, respectively. The learning rates for critic and policy model are set to $1e-4$ and $1e-5$, respectively. During inference, we set the temperature to 0 for deterministic generation. Consistent with previous works \cite{wang2025spa, feng2025group}, we set success rate as evaluation metric. All experiments are conducted on 8 H20 GPUs. To accelerate inference, we adopt the library of vllm and fastchat \cite{10.1145/3600006.3613165}. The prompt for behavior cloning, trajectory segmentation, and hindsight model training are shown in Table \ref{data-collection-prompts}-\ref{GPT-4o-prompt}.

\section{Baselines}
\label{appendix.c}
\textbf{Prompt Engineering Baselines}: To evaluate the agentic capabilities of powerful LLMs, the frozen LLMs (i.e., Llama3.2, GPT4o, Gemeni2.5pro) are adopted to execute the task in the three benchmarks. The results for GPT4o and Gemini2.5pro are from \citet{feng2025group}, while the remaining results are evaluated with the prompts shown in Table xxx. 

\noindent
\textbf{Behavior Cloning Baselines}: To evaluate the performance gain from offline samples, the backbone is performing continual training with the high-quality expert trajectories or the pairwise trajectories. The results for SFT and RFT are from \citet{wang2025spa}, while the DPO is reproduced with the open-sourced code \citet{wang-etal-2025-steca}.

\noindent
\textbf{Reinforcement Learning based Fine-tuning Baselines}: To evaluate the performance gain from interacting with the environment in real time, the SFT backbone is performing multi-turn agentic reinforcement learning. For the methods with outcome-based reward (e.g., PPO, GRPO), we set the final scalar trajectory outcome as the final reward of the trajectory. For the methods with process reward (e.g., SPA), we ensure that the sum of process rewards is equal to the final scalar trajectory outcome. The results come from \citet{wang2025spa} and \citet{feng2025group}.
\begin{table*}[t]
\centering
\scriptsize
\setlength{\tabcolsep}{2pt}
\renewcommand{\arraystretch}{1.2}

\definecolor{rowgray}{gray}{0.95}

\renewcommand{\tabularxcolumn}[1]{m{#1}}

\newcolumntype{Y}{>{\raggedright\arraybackslash}X}
\newcolumntype{C}{>{\centering\arraybackslash}c}

\begin{tabularx}{\textwidth}{@{} C @{\hspace{1em}} Y @{\hspace{1em}} C @{}}
\toprule
\textbf{Segment (turn)} &
\multicolumn{1}{c}{\textbf{Conversations}} &
$\mathbf{(\hat{R},\hat{z}, \hat{R}_{him})}$ \\
\midrule

\multicolumn{3}{@{}c}{\textbf{Alfworld Case: Put a clean apple $\rightarrow$ Microwave}} \\
\midrule

\rowcolor{rowgray}
\makecell[c]{\textbf{Segment1 (turn1)}\\[2pt] Initial task understanding  and \\ first action} &
\cellcolor{rowgray}
\textbf{Task Define}: You are in the middle of a room. Looking quickly around you, you see a cabinet 8, a cabinet 7, a cabinet 6, a cabinet 5, a cabinet 4, a cabinet 3, a cabinet 2, a cabinet 1, a coffeemachine 1, a countertop 2, a countertop 1, a fridge 1, a garbagecan 1, a microwave 1, a sinkbasin 1, a stoveburner 4, a stoveburner 3, a stoveburner 2, a stoveburner 1, and a toaster 1. \textbf{Your task is to:} put a clean apple in microwave. \par \vspace{2pt}
\textbf{Thought}: I need to put a clean apple in the microwave. First, I should locate an apple and ensure it's clean before I can use it. \textbf{Action}: go to garbagecan 1.
& $(0.030,\ 0.220,\ 0.041)$ \\

\multirow{2}{*}[-1.5em]{\makecell[c]{\textbf{Segment2 (turn2-3)}\\  Acquiring the apple}} &
\textbf{Observation}: On the garbagecan 1, you see a apple 2. \par \vspace{2pt}
\textcolor{blue}{\textbf{Thought}: I see an apple in the garbage can. I should pick it up and inspect it to see if it's clean or if it needs to be cleaned before use. \textbf{Action}: take apple 2 from garbagecan 1.}
& \multirow{2}{*}[-1.5em]{\makecell[c]{$(\textcolor{red}{0.092},\ \textbf{0.291},\ \textcolor{red}{\textbf{0.168}\uparrow})$}} \\

&
\textbf{Observation}: You pick up the apple 2 from the garbagecan 1. \par \vspace{2pt}
\textcolor{blue}{\textbf{Thought}: The apple is not clean, so I should wash it to ensure it's suitable for eating or use in the microwave. \textbf{Action}: go to sinkbasin 1.}
& \\

\cellcolor{rowgray} & 
\cellcolor{rowgray}
\textbf{Observation}: On the sinkbasin 1, you see nothing. \par \vspace{2pt}
\textbf{Thought}: I'm at the sink. Now I can clean the apple to make sure it's ready to be put in the microwave. \textbf{Action}: clean apple 2 with sinkbasin 1.
& \cellcolor{rowgray} \\

\cellcolor{rowgray}\multirow{-2}{*}[1em]{\makecell[c]{\textbf{Segment3 (turn4-5)}\\ Cleaning the apple}} &
\cellcolor{rowgray}
\textbf{Observation}: You clean the apple 2 using the sinkbasin 1. \par \vspace{2pt}
\textbf{Thought}: The apple is now clean. I need to find the microwave and open it to place the apple inside. \textbf{Action}: go to microwave 1.
& \cellcolor{rowgray}\multirow{-2}{*}[1.0em]{\makecell[c]{$(0.045,\ 0.240,\ 0.068)$}} \\ 


\makecell[c]{\textbf{Segment4 (turn6)}\\[2pt] Preparing the microwave} &
\textbf{Observation}: The microwave 1 is closed. \par \vspace{2pt}
\textbf{Thought}: The microwave is closed. I'll need to open it before I can put the apple inside. \textbf{Action}: open microwave 1.
& $(0.016,\ 0.110,\ 0.011)$ \\

\cellcolor{rowgray}
\makecell[c]{\textbf{Segment5 (turn7)}\\[2pt] Placing the clean apple in the \\ microwave} &
\cellcolor{rowgray}
\textbf{Observation}: You open the microwave 1. The microwave 1 is open. In it, you see a apple 1, and a potato 2. \par \vspace{2pt}
\textbf{Thought}: The microwave is open and already contains an apple and a potato. I should place the clean apple inside the microwave, ensuring it's in a suitable spot where it won't get crushed by the potato. \textbf{Action}: put apple 2 in/on microwave 1.
& \cellcolor{rowgray} $(0.818,\ 0.139,\ 0.712)$ \\

\midrule
\multicolumn{3}{@{}c}{\textbf{Virtualhome Case: Wash clothes}} \\
\midrule
\cellcolor{rowgray} 
\makecell[c]{\textbf{Segment1 (turn1)}\\ Task understanding and \\ locating the basket for clothes} &
\cellcolor{rowgray}
\textbf{Task Define}: The task is Wash clothes (I grab basket for clothes. I walk into the kitchen. I put clothes in washing machine. I turn on washing machine.). \par \vspace{2pt}
\cellcolor{rowgray}
\textbf{Thought}: To successfully complete the task of washing clothes, I need to first locate the basket for clothes, as it is essential for gathering the laundry before proceeding to the washing machine. \textbf{Action}: Find basket for clothes.
\cellcolor{rowgray}
& \cellcolor{rowgray}$(0.415,\ 0.125,\ 0.275)$ \\ 

\multirow{2}{*}[-1.5em]{\makecell[c]{\textbf{Segment2 (turn2-3)}\\  Grabbing the basket and \\ moving toward the washing \\ machine}} &
\textbf{Observation}: basket\_for\_clothes is CLOSED and CLEAN. \par \vspace{2pt}
\textcolor{blue}{\textbf{Thought}: I observed that the basket for clothes is closed, so my logical next step is to grab it in preparation to access the clothes that need washing. \textbf{Action}: Grab basket for clothes.}
& \multirow{2}{*}[-1.5em]{\makecell[c]{$(\textcolor{red}{0.209},\ \textbf{0.299},\ \textcolor{red}{\textbf{0.332}\uparrow})$}} \\

&
\textbf{Observation}: You pick up the apple 2 from the garbagecan 1. \par \vspace{2pt}
\textcolor{blue}{\textbf{Thought}: The action ``Walk to dining room'' was likely chosen because the task involves moving to a different location to proceed with the washing process, possibly to find the washing machine or to transport the clothes inside it. \textbf{Action}: Walk to dining room.}
& \\

\cellcolor{rowgray} & 
\cellcolor{rowgray}
\textbf{Observation}: basket\_for\_clothes is CLOSED and CLEAN. You have basket\_for\_clothes. \par \vspace{2pt}
\textbf{Thought}: The task requires me to wash clothes, but currently, I am holding a basket for clothes and standing in the dining room. I should locate the washing machine to proceed with the task. \textbf{Action}: Walk to washing machine.
& \cellcolor{rowgray} \\

\cellcolor{rowgray}\multirow{-2}{*}[2em]{\makecell[c]{\textbf{Segment3 (turn4-5)}\\ Reaching the washing \\ machine and preparing it}} &
\cellcolor{rowgray}
\textbf{Observation}: washing\_machine is OFF and CLOSED and CLEAN and PLUGGED\_IN, basket\_for\_clothes is CLOSED and CLEAN. You have basket\_for\_clothes. \par \vspace{2pt}
\textbf{Thought}: I have successfully reached the washing machine with the basket of clothes, and my next logical step is to open the washing machine so I can put the clothes in. \textbf{Action}: Open washing machine.
& \cellcolor{rowgray}\multirow{-2}{*}[1.0em]{\makecell[c]{$(0.135,\ 0.227,\ 0.164)$}} \\ 

\makecell[c]{\textbf{Segment4 (turn6)}\\ Placing the basket \\ on the washing machine} &
\textbf{Observation}: basket\_for\_clothes is CLOSED and CLEAN, washing\_machine is PLUGGED\_IN and OPEN and OFF and CLEAN. You have basket\_for\_clothes. \par \vspace{2pt}
\textbf{Thought}: I will place the basket for clothes near the washing machine so that I can easily access the clothes to load them into the machine. \textbf{Action}: Put basket for clothes on washing machine.
& $(0.113,\ 0.120,\ 0.073)$ \\

\cellcolor{rowgray} & 
\cellcolor{rowgray}
\textbf{Observation}: basket\_for\_clothes is CLOSED and CLEAN, washing\_machine is OPEN and OFF and CLEAN and PLUGGED\_IN, basket\_for\_clothes ON washing\_machine. \par \vspace{2pt}
\textbf{Thought}: I chose to close the washing machine as the next logical step in the task of washing clothes after putting them inside, to prepare for starting the wash cycle. \textbf{Action}: Close washing machine.
& \cellcolor{rowgray} \\

\cellcolor{rowgray}\multirow{-2}{*}[2em]{\makecell[c]{\textbf{Segment5 (turn7-8)}\\ Closing and turning \\ on the washing machine}} &
\cellcolor{rowgray}
\textbf{Observation}: basket\_for\_clothes is CLOSED and CLEAN, washing\_machine is CLOSED and PLUGGED\_IN and OFF and CLEAN, basket\_for\_clothes ON washing\_machine. \par \vspace{2pt}
\textbf{Thought}: Since I have already placed the basket for clothes on the washing machine and closed it, the next logical step is to turn on the washing machine to start the washing process. \textbf{Action}: Switch on washing machine.
& \cellcolor{rowgray}\multirow{-2}{*}[1.0em]{\makecell[c]{$(0.128,\ 0.229,\ 0.156)$}} \\

\bottomrule
\end{tabularx}

\caption{More case study on the Alfworld and Virtualhome dataset. The left shows segmented results of the trajectory, where Segment (turn) denotes consecutive turns contained in the segment. The middle shows the explored trajectory, where blue texts highlight the highest importance scores segment. The right shows the tuple of raw segmental process rewards, hindsight information modulated rewards, and segmental importance ratio scores.}
\label{case_study_appendix_1}
\end{table*}
\begin{table*}[t]
\centering
\scriptsize
\setlength{\tabcolsep}{2pt}
\renewcommand{\arraystretch}{1.2}

\definecolor{rowgray}{gray}{0.95}

\renewcommand{\tabularxcolumn}[1]{m{#1}}

\newcolumntype{Y}{>{\raggedright\arraybackslash}X}
\newcolumntype{C}{>{\centering\arraybackslash}c}

\begin{tabularx}{\textwidth}{@{} C @{\hspace{1em}} Y @{\hspace{1em}} C @{}}
\toprule
\textbf{Segment (turn)} &
\multicolumn{1}{c}{\textbf{Conversations}} &
$\mathbf{(\hat{R},\hat{z}, \hat{R}_{him})}$ \\
\midrule

\multicolumn{3}{@{}c}{\textbf{Alfworld Case: Put a clean soapbar $\rightarrow$ Cabinet}} \\
\midrule

\rowcolor{rowgray}
\makecell[c]{\textbf{Segment1 (turn1)}\\[2pt]  Understanding the task and \\ initiating a search for the soapbar} &
\textbf{Task Define}: You are in the middle of a room. Looking quickly around you, you see a cabinet 4, a cabinet 3, a cabinet 2, a cabinet 1, a countertop 1, a garbagecan 1, a handtowelholder 1, a sinkbasin 2, a sinkbasin 1, a toilet 1, a toiletpaperhanger 1, and a towelholder 1. \textbf{Your task is to:} put a clean soapbar in cabinet. \par \vspace{2pt}
\textbf{Thought}: The task is to put a clean soapbar in a cabinet. I should first look for a soapbar in the immediate vicinity. \textbf{Action}: go to toilet 1.
& $(0.027,\ 0.122,\ 0.022)$ \\

\multirow{-2}{*}[-0.3em]{\makecell[c]{\textbf{Segment2 (turn2)}\\ Identifying and taking \\ the soapbar}} &
\textbf{Observation}: On the toilet 1, you see a soapbar 1, a spraybottle 1, and a toiletpaper 1. \par \vspace{2pt}
\textbf{Thought}: I've found a soapbar on the toilet. I need to pick it up to clean it before placing it in the cabinet. \textbf{Action}: take soapbar 1 from toilet 1.
& \multirow{-2}{*}[-1.0em]{\makecell[c]{$(0.065,\ 0.146,\ 0.064)$}} \\ 



\cellcolor{rowgray} & 
\cellcolor{rowgray}
\textbf{Observation}: You pick up the soapbar 1 from the toilet 1. \par \vspace{2pt}
\textcolor{blue}{\textbf{Thought}: With the soapbar in hand, I should go to the sink to clean it since it's important to ensure cleanliness before storing or using it. \textbf{Action}: go to sinkbasin 1.}
& \cellcolor{rowgray} \\

\cellcolor{rowgray}\multirow{-2}{*}[2em]{\makecell[c]{\textbf{Segment3 (turn3-4)}\\ Preparing and cleaning \\ the soapbar}} &
\cellcolor{rowgray}
\textbf{Observation}: On the sinkbasin 1, you see nothing. \par \vspace{2pt}
\textcolor{blue}{\textbf{Thought}: I'm at the sink. Now I should clean the soapbar to make sure it's clean before placing it in the cabinet. \textbf{Action}: clean soapbar 1 with sinkbasin 1.}
& \cellcolor{rowgray}\multirow{-2}{*}[1em]{\makecell[c]{$(\textcolor{red}{0.063},\ \textbf{0.389},\ \textcolor{red}{\textbf{0.165}\uparrow})$}} \\ 

\multirow{2}{*}[-0.5em]{\makecell[c]{\textbf{Segment4 (turn5-6)}\\ Locating and accessing \\ the cabinet}} &
\textbf{Observation}: You clean the soapbar 1 using the sinkbasin 1. \par \vspace{2pt}
\textbf{Thought}: The soapbar is now clean. I need to find a cabinet to store the soapbar in. \textbf{Action}: go to cabinet 1.
& \multirow{2}{*}[-1em]{\makecell[c]{$(0.013,\ 0.214,\ 0.018)$}} \\

&
\textbf{Observation}: The cabinet 1 is closed. \par \vspace{2pt}
\textbf{Thought}: I've reached the cabinet. I need to open it to be able to put the soapbar inside. \textbf{Action}: open cabinet 1.
& \\

\rowcolor{rowgray}
\multirow{-2}{*}[-0.5em]{\makecell[c]{\textbf{Segment5 (turn7)}\\ Storing the clean \\ soapbar}} &
\textbf{Observation}: You open the cabinet 1. The cabinet 1 is open. In it, you see a cloth 2. \par \vspace{2pt}
\textbf{Thought}: The cabinet is open and I see there's space inside. It's a suitable place to store the clean soapbar. \textbf{Action}: put soapbar 1 in/on cabinet 1.
& \multirow{-2}{*}[-1.0em]{\makecell[c]{$(0.832,\ 0.129,\ 0.730)$}} \\ 
\midrule
\multicolumn{3}{@{}c}{\textbf{Alfworld Case: Cool and put some pot $\rightarrow$ Stoveburner}} \\
\midrule

\rowcolor{rowgray}
\makecell[c]{\textbf{Segment1 (turn1)}\\[2pt]  Interpreting the task and begining \\ exploration} &
\textbf{Task Define}: You are in the middle of a room. Looking quickly around you, you see a cabinet 20, a cabinet 19, a cabinet 18, a cabinet 17, a cabinet 16, a cabinet 15, a cabinet 14, a cabinet 13, a cabinet 12, a cabinet 11, a cabinet 10, a cabinet 9, a cabinet 8, a cabinet 7, a cabinet 6, a cabinet 5, a cabinet 4, a cabinet 3, a cabinet 2, a cabinet 1, a coffeemachine 1, a countertop 2, a countertop 1, a diningtable 2, a diningtable 1, a drawer 6, a drawer 5, a drawer 4, a drawer 3, a drawer 2, a drawer 1, a fridge 1, a garbagecan 1, a microwave 1, a sinkbasin 1, a stoveburner 4, a stoveburner 3, a stoveburner 2, a stoveburner 1, and a toaster 1. \textbf{Your task is to:} cool some pot and put it in stoveburner. \par \vspace{2pt}
\textbf{Thought}: I need to locate a pot to cool and then place it on a stoveburner. I'll start by checking the nearest appliances. \textbf{Action}: go to toaster 1.
& $(0.039,\ 0.209,\ 0.051)$ \\



&
\textbf{Observation}: On the toaster 1, you see nothing. \par \vspace{2pt}
\textbf{Thought}: The toaster is not where a pot would be, but I need to check all options systematically. \textbf{Action}: go to stoveburner 1.
&  \\

\multirow{2}{*}[3em]{\makecell[c]{\textbf{Segment2 (turn2-3)}\\ Identifying the target \\ object (pot)}} &
\textbf{Observation}: On the stoveburner 1, you see a pot 1. \par \vspace{2pt}
\textbf{Thought}: I've found a pot on stoveburner 1. I should pick it up so I can cool it as per the task. \textbf{Action}: take pot 1 from stoveburner 1.
& \multirow{-2}{*}[1em]{\makecell[c]{$(0.063,\ 0.249,\ 0.097)$}} \\ 

\rowcolor{rowgray}
\makecell[c]{\textbf{Segment3 (turn4-6)}\\[2pt] Picking up and cooling the pot with \\ the fridge} & 

\textbf{Observation}: You pick up the pot 1 from the stoveburner 1. \par \vspace{2pt}
\textcolor{blue}{\textbf{Thought}: With the pot in hand, I need to find a way to cool it down. The fridge is typically used for cooling items, so I'll head there. \textbf{Action}: go to fridge 1.} 


\textbf{Observation}: The fridge 1 is closed. \par \vspace{2pt}
\textcolor{blue}{\textbf{Thought}: The fridge is closed, but I can open it to cool the pot inside. \textbf{Action}: cool pot 1 with fridge 1.}


\textbf{Observation}: You cool the pot 1 using the fridge 1. \par \vspace{2pt}
\textcolor{blue}{\textbf{Thought}: Now that the pot is cooled, I need to place it back on a stoveburner. Stoveburner 1 is where I took it from, so it's the most convenient spot. \textbf{Action}:  go to stoveburner 1.}
& 
$(\textcolor{red}{0.058},\ \textbf{0.409},\ \textcolor{red}{\textbf{0.147}\uparrow})$ \\

\makecell[c]{\textbf{Segment5 (turn7)}\\ Placing the cooled pot \\ on the stoveburner.} &
\textbf{Observation}: On the stoveburner 1, you see nothing. \par \vspace{2pt}
\textbf{Thought}: I'm back at stoveburner 1, and it's clear. I can now place the cooled pot here to complete the task. \textbf{Action}: put pot 1 in/on stoveburner 1.
& $(0.841,\ 0.134,\ 0.705)$ \\

\bottomrule
\end{tabularx}

\caption{More case study on the Alfworld dataset. The left shows segmented results of the trajectory, where Segment (turn) denotes consecutive turns contained in the segment. The middle shows the explored trajectory, where blue texts highlight the highest importance scores segment. The right shows the tuple of raw segmental process rewards, hindsight information modulated rewards, and segmental importance ratio scores.}
\label{case_study_appendix_2}
\end{table*}

\begin{table*}[t]
    \footnotesize
    \begin{tabularx}{\textwidth}{lX}
        \toprule
        \textbf{Format} & \textbf{Prompt}\\
        \midrule
        \textbf{Alfworld} & \\
        \underline{Human} & {\ttfamily Interact with a household to solve a task. Imagine you are an intelligent agent in a household environment, and your target is to perform actions to complete the task goal. At the beginning of your interactions, you will be given a detailed description of the current environment and your goal to accomplish.\newline
        For each of your turns, you will be given the observation of the last turn. You should first think about the current condition and plan for your future actions, and then determine your action in this turn. Your output must strictly follow this format:''Thought: your thoughts.\newline
        Action: your next action''.\newline
        \newline
        The admissible actions are:\newline
        1. go to \{recep\}\newline
        2. task \{obj\} from \{recep\}\newline
        3. put \{obj\} in/on \{recep\}\newline
        4. open \{recep\}\newline
        5. close \{recep\}\newline
        6. toggle \{obj\} \{recep\}\newline
        7. clean \{obj\} with \{recep\}\newline
        8. heat \{obj\} with \{recep\}\newline
        9. cool \{obj\} with \{recep\}\newline
        where \{obj\} and \{recep\} correspond to objects and receptacles.\newline
        After your each turn, the environment will give you immediate feedback based on which you plan your next few steps. If the environment returns ''Nothing happened'', that means the previous action is invalid, and you should try other reasonable actions.\newline
        \newline
        Your response should use the following format:\newline
        \newline
        Thought: <your thoughts>\newline
        Action: <your next action>\newline
        \newline
        \textbf{Critical INSTRUCTIONS:}\newline
        1. Action Format: You must use the exact action formats provided above. NEVER simplify or modify them.\newline
        2. Grounding in History (previous actions and observations): When planning your future actions, you have to be strictly based on the history (previous actions and observations). Never assume anything not explicitly stated in the history.\newline
        3. Precise Task Decomposition: You should accurately understand and decompose the given task. Never overlook any details.} \\ \cmidrule(lr){2-2}
        \underline{GPT} & \texttt{OK, ......} \\
        \bottomrule
        \end{tabularx}
\caption{Prompt template for behavior cloning and trajectory collection in the AlfWorld dataset.}
\label{data-collection-prompts}
\centering
\end{table*}

\begin{table*}[t]
    \footnotesize
    \begin{tabularx}{\textwidth}{lX}
        \toprule
        \textbf{Format} & \textbf{Prompt}\\
        \midrule
        \textbf{VirtualHome} & \\
        \underline{Human} & {\ttfamily Interact with a household to solve a task. Imagine you are an intelligent agent in a household environment, and your target is to perform actions to complete the task goal. At the beginning of your interactions, you will be given a detailed description of the current environment and your goal to accomplish. \newline
        For each of your turns, you will be given the observation of the last turn. You should first think about the current condition and plan for your future actions, and then determine your action in this turn. Your output must strictly follow this format:''Thought: your thoughts. \newline
        Action: your next action''. \newline
        \newline
        The admissible actions are: \newline
        walk to \{obj\} \newline
        run to \{obj\} \newline
        grab \{obj\} \newline
        open \{obj\} \newline
        close \{obj\} \newline
        put \{obj\} on \{recep\} \newline
        put \{obj\} in \{recep\} \newline
        switch on \{obj\} \newline
        switch off \{obj\} \newline
        drink \{obj\} \newline
        look at \{obj\} \newline
        sit on \{obj\} \newline
        stand up \newline
        watch \{obj\} \newline
        wipe \{obj\} \newline
        type on \{obj\} \newline
        wash \{obj\} \newline
        cut \{obj\} \newline
        eat \{obj\} \newline
        sleep \newline
        wake up \newline
        plug in \{obj\} \newline
        plug out \{obj\} \newline
        pour \{obj\} into \{recep\} \newline
        move \{obj\} \newline
        release \newline
        turn to \{obj\} \newline
        where \{obj\} and \{recep\} correspond to objects and receptacles. \newline
        After your each turn, the environment will give you immediate feedback based on which you plan your next few steps. If the environment returns ''Nothing happened'', that means the previous action is invalid, and you should try other reasonable actions. \newline
        \newline
        Your response should use the following format: \newline
        \newline
        Thought: <your thoughts> \newline
        Action: <your next action> \newline
        \newline
        \textbf{Critical INSTRUCTIONS:}\newline
        1. Action Format: You must use the exact action formats provided above. NEVER simplify or modify them. \newline
        2. Grounding in History (previous actions and observations): When planning your future actions, you have to be strictly based on the history (previous actions and observations). Never assume anything not explicitly stated in the history. \newline
        3. Precise Task Decomposition: You should accurately understand and decompose the given task. Never overlook any details.} \\ \cmidrule(lr){2-2}
        \underline{GPT} & \texttt{OK, ......} \\
        \bottomrule
    \end{tabularx}
\caption{Prompt template for behavior cloning and trajectory collection in the Virtualhome dataset.}
\label{data-collection-prompts}
\centering
\end{table*}

\begin{table*}[t]
    \footnotesize
    \begin{tabularx}{\textwidth}{lX}
        \toprule
        \textbf{Format} & \textbf{Prompt}\\
        \midrule
        \textbf{Webshop} & \\
        \underline{Human} & {\ttfamily You are web shopping. I will give you instructions about what to do. You have to follow the instructions. Every round I will give you an observation and a list of admissible actions, you have to respond an action based on the state and instruction.\newline
        You can use search action if search is admissible. You can click one of the buttons in clickables. Your output must strictly follow this format:''Thought: your thoughts.\newline
        Action: your next action''.\newline
        \newline
        The admissible actions should be of the following structure:\newline
        search[keywords]\newline
        click[value]\newline
        where [keywords] in search are up to you, but the [value] in click must be a value in the list of admissible actions.\newline
        If the action is not valid, perform nothing. Remember that your keywords in search should be carefully designed.\newline
        \newline
        Your response should use the following format:\newline
        \newline
        Thought: <your thoughts>\newline
        Action: <your next action>\newline
        \newline
        \textbf{Critical INSTRUCTIONS:}\newline
        1. Action Format: You must use the exact action formats provided above. NEVER simplify or modify them.\newline
        2. Grounding in History (previous actions and observations): When planning your future actions, you have to be strictly based on the history (previous actions and observations). Never assume anything not explicitly stated in the history.\newline
        3. Precise Task Decomposition: You should accurately understand and decompose the given task. Never overlook any details.} \\ \cmidrule(lr){2-2}
        \underline{GPT} & \texttt{OK, ......} \\
        \bottomrule
        \end{tabularx}
\caption{Prompt template for behavior cloning and trajectory collection in the Webshop dataset.}
\label{data-collection-prompts}
\centering
\end{table*}

\begin{table*}[t]
\footnotesize
\begin{tabularx}{\textwidth}{lX}
    \toprule
    \textbf{Role} & \textbf{Content} \\
    \midrule
    \textbf{Human} & {\ttfamily Imagine you are a trajectory expert of household agents. You are given a completed exploration trajectory of a household agent in the environment for a certain task, but one of the interactive turns of the household agent is lost (represented by the underline \textquotesingle\_\_\_\textquotesingle). Your goal is to predict the lost interactive turn, like the task of cloze. At the begining of the given trajectory, you can see a detailed description of the environment in which the household agent is located and the task to be accomplished. Then you can see all interactive turns except for the lost one. Each interative turn contains the \textquotesingle Observation\textquotesingle\ returned by the environment (Note: if the envrionment outputs "Nothing happened", it means the previous action is invalid), the \textquotesingle Thought\textquotesingle\ and \textquotesingle Action\textquotesingle\ output by the household agent. At the end, you can see the scalar reward for the given trajectory, which reflects how well the trajectory accomplishes the task (\textquotesingle 0.00\textquotesingle\ and \textquotesingle 1.00\textquotesingle\ represent complete failure and success of the task, respectively). When predicting the lost interactive turn, you should comprehensively consider the visible context before and after the lost interactive turn as well as the final scalar reward. You are only required to output the lost \textquotesingle Thought\textquotesingle\ and \textquotesingle Action\textquotesingle. Your output must strictly follow this format:"Thought: your thoughts.\newline Action: your next action". \newline
    \newline
    The available actions are: \newline
    1. go to \{recep\} \newline
    2. task \{obj\} from \{recep\} \newline
    3. put \{obj\} in/on \{recep\} \newline
    4. open \{recep\} \newline
    5. close \{recep\} \newline
    6. toggle \{obj\} \{recep\} \newline
    7. clean \{obj\} with \{recep\} \newline
    8. heat \{obj\} with \{recep\} \newline
    9. cool \{obj\} with \{recep\} \newline
    where \{obj\} and \{recep\} correspond to objects and receptacles. \newline
    \newline
    Your response should only include the lost turn in the following format: \newline
    \newline
    Thought: <your thoughts> \newline
    Action: <your next action>} \\
    \midrule
    \textbf{GPT} & {\ttfamily OK} \\
    \midrule
    \textbf{Human} & {\ttfamily You are in the middle of a room. Looking quickly around you, you see a armchair 1, a bed 1, a diningtable 1, a drawer 2, a drawer 1, a garbagecan 1, a sidetable 2, and a sidetable 1. \newline
    \newline
    Your task is to: find two pillow and put them in armchair.} \\
    \midrule
    \textbf{GPT} & {\ttfamily \_\_\_} \\
    \midrule
    \textbf{Human} & {\ttfamily Observation: On the sidetable 1, you see a cellphone 3, and a keychain 2.} \\
    \midrule
    \textbf{GPT} & {\ttfamily Thought: There are no pillows on sidetable 1, only a cellphone and a keychain. I should check the other side table. \newline
    Action: go to sidetable 2} \\
    \midrule
    \textbf{...} & \textbf{... ...} \\
    \midrule
    \textbf{Human} & {\ttfamily The scalar reward for the above trajectory is: 1.0 \newline
    Predict the lost interactive turn, like the task of cloze} \\
    \midrule
    \midrule
    \textbf{Target} & {\ttfamily Thought: I need to find two pillows to place on the armchair. I\textquotesingle ll start by searching the nearest surfaces where pillows are likely to be found, such as side tables. \newline
    Action: go to sidetable 1} \\
    \bottomrule
\end{tabularx}
\caption{Prompt template for hindsight model training in the Alfworld \& Virtualhome dataset.}
\label{data-collection-prompts}
\centering
\end{table*}

\begin{table*}[t]
\footnotesize
\begin{tabularx}{\textwidth}{lX}
    \toprule
    \textbf{Role} & \textbf{Content} \\
    \midrule
    \textbf{Human} & {\ttfamily Imagine you are a trajectory expert of webshop agents. You are given a completed exploration trajectory of webshop agent in the web for a certain task, but one of the interactive turns of the webshop agent is lost (represented by the underline "\_\_\_"). Your goal is to predict the lost interactive turn, like the task of cloze. At the beginning of the given trajectory, you can see a detailed description of the task that the webshop agent has to accomplish. Then you can see all interactive turns except for the lost one. Each interative turn contains the "Observation" returned by the web, the "Thought" and "Action" output by the webshop agent. At the end, you can see the scalar reward for the given trajectory, which reflects how well the trajectory accomplishes the task ("0.00" and "1.00" represent complete failure and success of the task, respectively). When predicting the lost interactive turn, you should comprehensively consider the visible context before and after the lost interactive turn, as well as the final scalar reward. Your prediction for the lost interactive turn should include only the "Thought" and "Action" fields, in the format below: \newline
    \newline
    Thought: <your thoughts> \newline
    Action: <your next action>} \\
    \midrule
    \textbf{GPT} & {\ttfamily OK} \\
    \midrule
    \textbf{Human} & {\ttfamily WebShop [SEP] Instruction: [SEP] i\textquotesingle m looking for native american indian dream catcher feathers talisman, and price lower than 50.00 dollars [SEP] Search} \\
    \midrule
    \textbf{GPT} & {\ttfamily \_\_\_} \\
    \midrule
    \textbf{Human} & {\ttfamily Observation: \newline
    Instruction: [SEP] i\textquotesingle m looking for native american indian dream catcher feathers talisman, and price lower than 60.00 dollars [SEP] Back to Search [SEP] Page 1 (Total results: 50) [SEP] Next > [SEP] B07TGHZH66 [SEP] BlessLiving Wolf Dream Catcher Comforter Set Native American Wolf Comforter King Size Bed Set Indian Wolf Bedding Sets King, Brown Gold, 3 Piece, 1 Quilt 2 Shams [SEP] \$69.99 [SEP] B09Q61XL96 [SEP] Native American Dream Catcher Beads Women Chiffon Short Sarongs Cover Ups Bikini Wrap Scarf Black [SEP] \$7.77 [SEP] B08ZXQW35Z [SEP] Bohemian Dream Catcher Printed Shade Durm Shape Lampshade for Table Lamp and Floor Light, Home Decor, 7.87 R x 7.08 H inches [SEP] \$32.99 [SEP] B095VX97GN [SEP] Native American Indian Dream Catcher Feathers Talisman 3 PCs Set, 1 Women Travel Cosmetic Bag, 1 Hanging Toiletry Wash bag, 1 Pencil case, Portable Makeup Organizer Brush Storage Case [SEP] \$33.99 [SEP] ...} \\
    \midrule
    \textbf{GPT} & {\ttfamily Thought: I think I should click on the product that matches the description of a Native American Indian dream catcher feathers talisman and is priced below \$60.00. \newline
    Action: click[B095VX97GN]} \\
    \midrule
    \textbf{...} & \textbf{... ...} \\
    \midrule
    \textbf{Human} & {\ttfamily The scalar reward for the above trajectory is: 1.0 \newline
    Predict the lost interactive turn, like the task of cloze} \\
    \midrule
    \midrule
    \textbf{Target} & {\ttfamily Thought: I think I should use the search bar to look for the product I need. \newline
    Action: search[inner connection to spirit and talisman native american indian dream catcher]} \\
    \bottomrule
\end{tabularx}
\caption{Prompt template for hindsight model training in the Webshop dataset.}
\label{data-collection-webshop-388}
\centering
\end{table*}

\begin{table*}[t]
    \footnotesize
    \begin{tabularx}{\textwidth}{p{2.2cm}X} 
        \toprule
        \textbf{Dataset} & \textbf{Prompt}\\
        \toprule
        \textbf{Alfworld \& Virtualhome} & 
        \texttt{You are an expert at segmenting multi-turn interactive trajectories between intelligent agents and household environments into meaningful stages.\newline
        You are given a multi-turn interactive trajectory where each round contains a 'user' (the environment: initial task or feedback/observation) and an 'assistant' (the agent: reasoning 'Thought' and chosen 'Action') interaction.\newline
        Your task is to segment this trajectory into several consecutive segments, where each segment represents a distinct stage or subgoal in the agent's progress toward the specified task.\newline
        \newline
        \textbf{Critical INSTRUCTIONS:}\newline
        \newline
        1. Segmentation criteria:\newline
        $\bullet$ Each segment should reflect a clear change in the agent's high-level intention, intermediate sub-goal, strategy shift, or a significant event in the environment (for example, a new search target, a new object found, switching from searching for an object to manipulating it, picking up a crucial object, or a clear environmental cue causes a shift in plan).\newline
        $\bullet$ Segments can span several turns or just one, but adjacent segments must differ in agent intent or environmental context.\newline
        $\bullet$ Do not make segments too coarse: Prefer more segments with fewer turns each, unless the trajectory truly supports only a few distinct stages. Too few segments (fewer than \{\} turns) suggest your segmentation may be coarse.\newline
        2. Output requirements:\newline
        $\bullet$ For each segment you identify, briefly explain the stage transition or boundary in 1-2 sentences; describe what defines the start/end of that segment.\newline
        $\bullet$ At the end, return all index ranges (1-based) for each segment clearly in this format: [(1, 3), (4, 6), (7), (8, 13), (14, 20)], the index ranges should be consecutive.\newline
        $\bullet$ Ensure the total number of segments is at least 4 whenever possible, unless the trajectory is exceptionally short.\newline
        \newline
        Example:\newline
        \newline
        For a trajectory of 8 turns, you may segment as follows:\newline
        (1, 3): Agent interprets the task and starts searching.\newline
        (4, 6): Agent finds possible location and investigates.\newline
        (7): Agent acquires object.\newline
        (8, 13): Agent manipulates the object.\newline
        (14, 20): Agent arrives at the destination and puts on the obejct to compelete the task.\newline
        and then return the index ranges at the end of your response in the format: ''[(1, 3), (4, 6), (7), (8, 13), (14, 20)]''\newline
        \newline
        Now, analyze the following provided multi-turn interactive trajectory (wrapped in **********) step by step, explain transitions, return the index ranges for each segment in the Square brackets [] at the end of your response in the format: 'The segmented results are: [fill the segmented results here]'} \\
        \bottomrule
    \end{tabularx}
\caption{Prompt templates for GPT-4o to perform trajectory segmentation across Alfworld \& Virtualhome datasets.}
\label{GPT-4o-prompt}
\centering
\end{table*}

\begin{table*}[t]
    \footnotesize
    \begin{tabularx}{\textwidth}{p{2.2cm}X} 
        \toprule
        \textbf{Dataset} & \textbf{Prompt}\\
        \toprule
        \textbf{Webshop} & 
        \texttt{You are an expert at segmenting multi-turn interactive trajectories between intelligent agents and webshopping environments into meaningful stages.\newline
        You are given a multi-turn interactive trajectory where each round contains a 'user' (the environment: initial task or feedback/observation) and an 'assistant' (the agent: reasoning 'Thought' and chosen 'Action') interaction.\newline
        Your task is to segment this trajectory into several consecutive segments, where each segment represents a distinct stage or subgoal in the agent's progress toward the specified task.\newline
        \newline
        \textbf{Critical INSTRUCTIONS:}\newline
        \newline
        1. Segmentation criteria:\newline
        $\bullet$ Each segment should reflect a clear change in the agent's high-level intention, intermediate sub-goal, strategy shift, or a significant event in the environment (for example, a new search target, switching from searching to clicking, deciding to buy or perform another action, or a clear environmental cue causes a shift in plan).\newline
        $\bullet$ Segments can span several turns or just one, but adjacent segments must differ in agent intent or environmental context.\newline
        $\bullet$ Do not make segments too coarse: Prefer more segments with fewer turns each, unless the trajectory truly supports only a few distinct stages. Too few segments (fewer than \{\} turns) suggest your segmentation may be coarse.\newline
        2. Output requirements:\newline
        $\bullet$ For each segment you identify, briefly explain the stage transition or boundary in 1-2 sentences; describe what defines the start/end of that segment.\newline
        $\bullet$ At the end, return all index ranges (1-based) for each segment clearly in this format: [(1, 3), (4, 6), (7), (8, 10)]. the index ranges should be consecutive.\newline
        $\bullet$ Ensure the total number of segments is at least 3 whenever possible, unless the trajectory is exceptionally short.\newline
        \newline
        Example:\newline
        \newline
        For a trajectory of 8 turns, you may segment as follows:\newline
        (1, 3): Agent interprets the task and starts searching.\newline
        (4, 6): Agent searchs relevant objects in the webshop.\newline
        (7): Agent clicks a corresponding button.\newline
        (8, 10): Agent finds the target and clicks the 'Buy Now' button to complete the task.\newline
        and then return the index ranges at the end of your response in the format: ''[(1, 3), (4, 6), (7), (8, 10)]''\newline
        \newline
        Now, analyze the following provided multi-turn interactive trajectory (wrapped in **********) step by step, explain transitions, return the index ranges for each segment in the Square brackets [] at the end of your response in the format: 'The segmented results are: [fill the segmented results here]'} \\
        \bottomrule
    \end{tabularx}
\caption{Prompt templates for GPT-4o to perform trajectory segmentation in the Webshop dataset.}
\label{GPT-4o-prompt}
\centering
\end{table*}

\end{document}